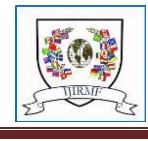

# CNN-LSTM Hybrid Deep Learning Model for Remaining Useful Life Estimation

**[1]Muthukumar G.,   [2]Jyosna Philip**
[1] Technical Officer C, Indira Gandhi Centre for Atomic Research (IGCAR), Kalpakkam, Tamilnadu
Email: muthuganeshece@gmail.com / muthukumarg@igcar.gov.in
[2] Student, M.Sc. (Data Science), Christ (deemed to be) University, Pune, India.
Email: philipjyosna02@gmail.com

**Abstract:**  *Remaining Useful Life (RUL) of a component or a system is defined as the length from the current time to the end of the useful life. Accurate RUL estimation plays a crucial role in Predictive Maintenance applications. Data driven approaches for RUL estimation use sensor data and operational data to estimate RUL. Traditional regression methods, both linear and non-linear, have struggled to achieve high accuracy in this domain. Although Multilayer Perceptron (MLP) has been applied to predict RUL, it cannot learn salient features automatically, because of its network structure. While Convolutional Neural Networks (CNNs) have shown improved accuracy, they often overlook the sequential nature of the data, relying instead on features derived from sliding windows. Since RUL prediction inherently involves multivariate time series analysis, robust sequence learning is essential. In this work, we propose a hybrid approach combining Convolutional Neural Networks with Long Short-Term Memory (LSTM) networks for RUL estimation. Although CNN-based LSTM models have been applied to sequence prediction tasks in financial forecasting, this is the first attempt to adopt this approach for RUL estimation in prognostics. In this approach, CNN is first employed to efficiently extract features from the data, followed by LSTM, which uses these extracted features to predict RUL. This method effectively leverages sensor sequence information, uncovering hidden patterns within the data, even under multiple operating conditions and fault scenarios. For comparative purpose, we also evaluate the performance of various machine learning algorithms including Gradient Boosting, MLP, CNN, LSTM and Random Forest on the NASA CMAPSS dataset, which includes sensor data linked to the RUL of various jet engines. Our results demonstrate that the hybrid CNN-LSTM model achieves the highest accuracy, offering a superior R² score compared to the other methods.*

| Model | RMSE | R2 Score |
|---|---|---|
| Linear Regression | 43.18 | 0.46 |
| Random Forest | 6.68 | 0.42 |
| XG Boost | 17.35 | 0.65 |
| Multilayer Perceptron | 4.51 | 0.52 |
| CNN | 16.82 | 0.79 |
| LSTM | 15.93 | 0.75 |
| **CNN + LSTM** | **13.34** | **0.86** |

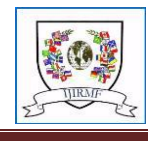

## 1.  INTRODUCTION

Remaining Useful Life (RUL) of a component or a system is defined as the length from the current time to the end of the useful life [2]. Accurate RUL estimation plays a crucial role in Predictive Maintenance applications. If we can accurately predict when an engine will fail, then we can make informed maintenance decision in advance to avoid disasters, reduce the maintenance cost, as well as streamline operational activities, aligning with the principles of industry 4.0 [1, 10]. In general, two types of methodologies are used for RUL estimation: model-based approaches and data-driven approaches. Model-based methodologies build physical failure models for degradation, such as crack, wear, corrosion, etc [1]. Physical models are very useful to solve RUL problem in use-cases where there is no enough failure data available. In such cases, we can induce failures within physical models, augment the actual data with physical model failure data and learn models for RUL estimation. However, such physical failure models are very complex and difficult to build, and physical models for many components do not exist. On the other hand, data-driven methods that employ sensor and operational status data to estimate RUL are more advantageous and economical for equipment with a sufficient number of failures.

In this paper, we conducted a thorough analysis of the data driven approach for RUL estimation. Traditional data driven approach utilizes both linear and non-linear regression techniques to estimate RUL, however they have difficulty achieving high accuracy in this field. Although Multilayer Perceptron (MLP) has been applied to predict RUL, it cannot learn salient features automatically, because of its network structure. However, it is extremely challenging, if not impossible, to accurately predict RUL without a good feature representation method. It is thus highly desirable to develop a systematic feature representation approach to effectively characterize the nature of signals related to the prognostic tasks.

Recently deep learning models are highly popular due to its ability to learn automatically the hierarchical feature representation from raw data. The deep learning architecture contains sequence of layers, each of which applies a non-linear transformation on the outputs of the previous layer. This allows the data to be represented by a hierarchy of features with varying levels of detail. The widely known deep learning models are Convolutional neural network (CNN), Recurrent Neural Network (RNN), Auto-encoders and Transformers.

Convolutional neural network adapted from deep learning architecture uses different processing units such as convolution, pooling, activation etc., to effectively capture local features from the global data [27]. The deep architecture allows multiple layers of these units to be stacked, enabling the model to identify signal characteristics at different scales. Therefore, the features extracted by CNN are task dependent and non-handcrafted. Moreover, these features offer more predictive power, as the CNN is trained under the supervision of target values.

Recurrent Neural Network, a class of deep learning architectures is well suited for modelling time sequence data [25]. However, RNN is known to suffer from long-term temporal dependency problem, as the gradients propagated over multiple stages tend to either vanish or explode. Long Short-Term Memory Network (LSTM) is a form of RNN network for sequence learning tasks [5, 21] and has achieved remarkable success on speech recognition and machine translation. LSTM addresses the long-term time dependency problem of RNN by controlling information flow using input gate, forget gate and output gate. Long term memory retention is essentially their default mode of operation [23].

While Convolutional Neural Networks have shown improved accuracy, they often overlook the sequential nature of the data, relying instead on features derived from sliding windows. Since RUL prediction inherently involves multivariate time series analysis, robust sequence learning is essential.

Therefore, a novel hybrid architecture combining CNN with LSTM networks for RUL estimation is developed in this paper. Although CNN-based LSTM models have been applied to sequence prediction tasks in financial forecasting, this is the first attempt to adopt this approach for RUL estimation in prognostics. In the proposed architecture for RUL estimation, one dimensional convolutional filters in the initial layer are applied to all the sensor data at each time stamp followed by LSTM layers applied temporally over the time series and the final neural network regression layer employs squared error loss function. In this approach, the CNN layer efficiently extracts features from the data, followed by LSTM layer, which uses these extracted features to predict RUL. This method effectively leverages sensor sequence information, uncovering hidden patterns within the data, even under multiple operating conditions and fault scenarios.

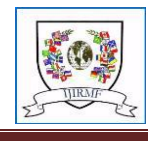

Data-driven approaches involve several key steps, including data collection, data wrangling, exploratory data analysis, feature engineering, and subsequent model preparation and evaluation. In the experiments, the proposed CNN with LSTM based approach for RUL estimation is compared with existing regression-based approaches in the CMAPSS public data set. The results clearly demonstrates that the proposed approach accurately predicts RUL than existing approaches significantly.

The following sections of this article are organized as follows: Section 2 explains the background and related work. Section 3 outlines the CNN LSTM hybrid Deep learning architecture. Section 4 describes the methodology, and Section 5 discusses the experimental results. Finally, Section 6 addresses the future work and conclusions.

## 2.  Related Work :

In this section we primarily focus on regression-based machine learning approaches for RUL estimation. There exist two main categories of machine learning-based techniques, the first one is supervised approaches where the failure information exists in the dataset and the second one is unsupervised approaches, where there is only process information, and no failure-related information exists [30]. Supervised machine learning methods have been increasingly applied for RUL estimation in the various areas such as medical devices [12], automated teller machines (ATM) [13], electric propulsion systems [14, 29], rolling-element bearings [15], computer workstations [16], automobiles [28] and industrial machines [29].

The existing algorithms in the literature for RUL estimation are either based on multivariate time series analysis or damage progression analysis [3, 18, 19, 20, 26]. Many approaches utilize conventional machine learning models such as support vector machines (SVM) [17] and decision tree-based models [12, 13]. An important benefit of these models is their interpretability, as they help identify key factors contributing to machine breakdowns.

In [4], a deep convolutional neural network model is used to estimate RUL. In the proposed architecture, convolution filters in the initial layer are two dimensional which is applied along the temporal dimension over the multi-channel sensor data and final neural network regression layer employs squared error loss function to incorporate automated feature learning from raw sensor signals in a systematic way. Li et al. [32] developed a multi-scale deep convolutional neural network (MS-DCNN) and used the min-max normalization with the MS-DCNN algorithm for RUL prediction. They compared the performance of their model with other state-of-the-art models and showed that the new model provides promising results on the NASA C-MAPSS dataset [26].

In [5], LSTM network model is utilised for the accurate RUL estimation. Due to the inherent sequential nature of sensor data, LSTM is well suited for RUL estimation. In this approach, multiple layers of LSTM cells in combination with feed forward layers to uncover hidden patterns in sensor data at various stages of degradation.

A set of turbo engine run-to-failure datasets is provided by NASA [26, 31], and the data is used in many research papers to predict the RUL. Ramasso and Saxena [24] published a survey on prognostic methods used for the NASA turbo engine datasets and divided the prognostic approaches into three categories. The first category is the use of functional mappings between the set of inputs and RUL. For the first category, they reported that the dominant underlying machine learning algorithm is artificial neural networks (ANNs) [22]. The second category of techniques is the functional mapping between the health index and RUL. The third category is similarity-based matching techniques. Benchmarking of prognostic methods has been conducted on the NASA turbo engine dataset, and it was shown that most of the studies use a health index to map between input features and the RUL.

In several studies, indirect measurements of RUL are used in place of direct observations. Because of this, estimating the RUL frequently involves the use of the health index (HI) concept [34]. Rather than directly predicting the RUL, a machine learning model is trained to forecast the HI of a turbo engine for each cycle. Ziqiu et al. [33] used the notion of HI and the Multilayer Perceptron (MLP) machine learning method in combination with feature normalization, Principal Component Analysis (PCA), and feature selection approaches to estimate the RUL. They also used the NASA CMAPSS dataset to assess the performance of their model.

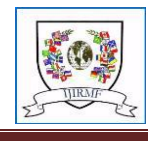

In our study, we adopt to the supervised machine learning approach for the estimation of RUL using a deep CNN combined with LSTM architecture. In consistent with other models, we thoroughly investigate the effectiveness of this innovative model and compare its performance against other established machine learning algorithms using the NASA turbo engine datasets.

### 3.   CNN LSTM Hybrid Deep Learning Architecture for RUL Estimation

This section presents the proposed architecture of CNN LSTM hybrid deep learning model for RUL estimation. CNN have great potential to identify the various salient patterns of sensor signals. However, in RUL estimation we confront with multiple channels of time series signals, in which the traditional CNN cannot be used directly. Hence, we hybrid the LSTM network to capture temporal dimension. The LSTM module receives the output from the CNN module and processes it sequentially to capture long-term dependencies and temporal patterns. It uses multiple layers of LSTM cells in combination with standard feed forward layers to discover hidden patterns from sensor data. The overview block diagram of the CNN LSTM architecture is shown in the Fig. 1.

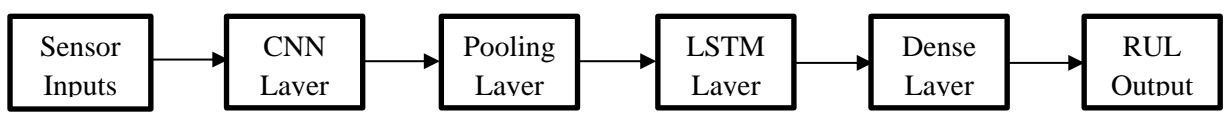

*Fig. 1: Block Diagram of proposed CNN LSTM hybrid architecture*

### 3.1.  Convolutional Neural Network

In our work, we have limited the architecture to a single CNN layer, which consists of one convolution layer followed by a pooling layer. This simplified structure is sufficient for this task, which helps to reduce computational complexity. In the convolution layer, the input sensor data at a specific time instant is processed using one-dimensional convolutional kernels. We apply 64 one dimensional convolution filters and relu activation function. In the pooling layer, we use one dimensional max pooling without overlapping, where the input feature maps are divided into non-overlapping regions, and for each sub-region, the maximum value is taken as the output.

### 3.2.  LSTM Model

LSTM cell uses memory cells and three gates: the forget gate determines what information to discard, the input gate controls new information added, and the output gate manages how much of the cell state is used for output as shown in Fig. 2. This structure allows LSTMs to effectively retain important information over time, making them suitable for tasks like time-series prediction and sequence modeling. Hence, we split the dataset into multiple fixed length time series data of length 30 and applied to the model.

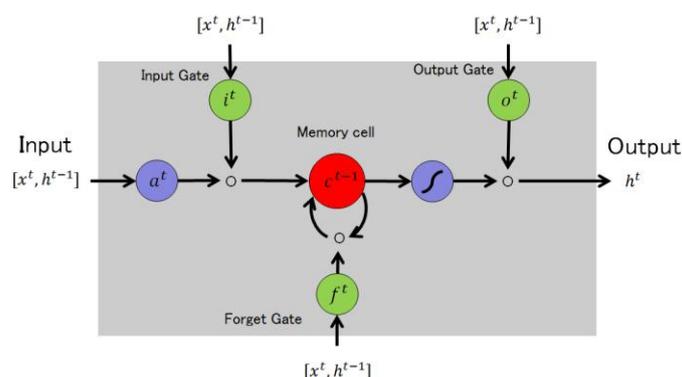

*Fig. 2: LSTM Cell*

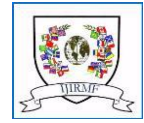

### 3.3. CNN LSTM Architecture

The CNN module processes the input data and extracts relevant spatial features through convolutional layers. These layers employ filters to capture local patterns within the input sequences, enabling the model to learn hierarchical representations of the data. The LSTM module receives the output from the CNN module and processes it sequentially to capture long-term dependencies and temporal patterns. In the proposed CNN LSTM hybrid architecture shown in Fig. 3, leverages the strengths of both CNNs and LSTMs to capture spatial patterns and long-term dependencies in sequential data. We combined the 1D convolution layer followed by pooling layer with multiple LSTM and fully connected layers. CNN takes the input sensor data at each time step and its output is passed to LSTM layer. The LSTM output is then fed into fully connected layers, and the final regression layer predicts the RUL.

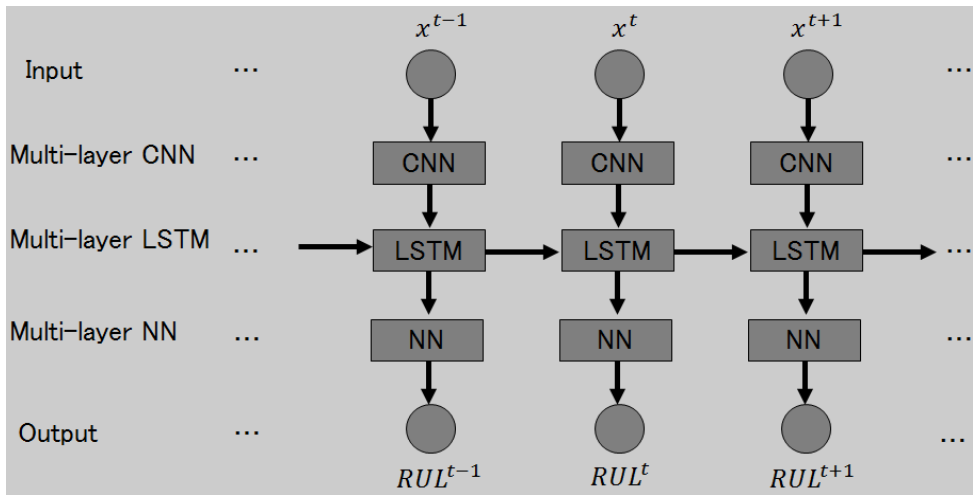

*Fig. 3: CNN LSTM Hybrid Architecture Model*

### 3.4. Model Evaluation

In order to evaluate the performance of a RUL estimation model on the test data, *Root Mean Square Error* (RMSE) Eq. (1), gives equal penalty weights to the model when the estimated RUL is smaller than true RUL and when the estimated RUL is larger than true RUL.

$$\text{RMSD} = \sqrt{\frac{1}{n}\sum_{i=1}^{n}(X_i - x_0)^2}. \tag{1}$$

Eq. (2) is *R² Score*, which is also widely used as an evaluation metric for the estimation of RUL. It represents the proportion of variance in the target variable that can be explained by the independent variables in the model. Mathematically, the R² score is defined as:

$$R^2 = 1 - \frac{SS_{\text{res}}}{SS_{\text{tot}}} \tag{2}$$

Where $SS_{\text{res}}$ is the sum of squared residuals (the differences between the observed and predicted values), and $SS_{\text{tot}}$ is the total sum of squares (the differences between the observed values and their mean). R² score of 1 indicates that the model perfectly explains all the variability in the target variable and an R² score of 0 suggests that the model fails to explain any variability and performs no better than a simple mean-based prediction.

In literature, *scoring* function is given to measure the quality of the models [4, 5]. Eq. (3) shows the definition of scoring function.

$$S = \begin{cases} \sum_{i=1}^{n}(e^{-\frac{h_i}{13}} - 1), \text{when } h_i < 0 \\ \sum_{i=1}^{n}(e^{\frac{h_i}{10}} - 1), \text{when } h_i \geq 0, \end{cases} \tag{3}$$

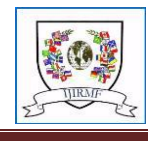

Where n is total number of samples in the test set, $h_i = RUL_{est,i} - RUL_i$, $RUL_i$ is true RUL for the test sample 'i'. Eq. (3) gives different penalty when the model underestimates RUL and when the model overestimates RUL. If estimated RUL is less than the true RUL, the penalty is smaller, because there is still time to conduct maintenance and it will not cause significant system failure. If estimated RUL is larger than true RUL, the penalty is larger, because under such estimation, the maintenance will be scheduled later than the required time and it may cause system failure.

## 4.  Methodology :

This section outlines the key steps taken before the modeling phase. It starts with preparing features and target variables, followed by data analysis to explore relationships between variables using visualizations. Data pre-processing is then discussed, including filtering and normalization. Feature engineering is applied to create or modify features for better predictive power. Finally, techniques like Principal Component Analysis (PCA) and feature selection are used to remove redundant or irrelevant features.

### 4.1. Data Preparation

The dataset comprises multiple multivariate time series sensor data, which is divided into training and test subsets. Each time series is from a different engine i.e., the data can be considered to be from a fleet of engines of the same type. Every engine has varying degrees of initial wear and manufacturing variance in the beginning. This variation and wear are regarded as normal and do not indicate a malfunction. The data is contaminated with sensor noise. The engine is operating normally at the start of each time series, and develops a fault at some point during the series. In the training set, the fault grows in magnitude until system failure. In the test set, the time series ends some time prior to system failure. The objective is to predict the number of remaining operational cycles before failure in the test set, or in other words the number of cycles the engine will continue to operate after the last recorded cycle. In addition to that a vector of true RUL for the test set is also provided in the dataset.

*CMAPSS Dataset*

The dataset contains 26 numerical features, where each row represents a snapshot of data collected in a single operational cycle across 100 different engines. Sensor data is recorded during each engine power cycle and is gathered for one hundred distinct engines. Engine ID, Time in Cycles, Settings 1, 2, 3, and the remaining columns all contain sensor data. There is no formal definition of the specifics of any sensor data.

*Target RUL*

The target variable is not explicitly provided in the dataset, instead it is calculated by subtracting the number of cycles from the engine's maximum cycle. This leads to a new column labelled "Remaining Cycles," which becomes the target variable and it represents the Remaining Useful Life of the engine. The health of a system generally degrades linearly along with time. In practical applications, a component's degradation is minimal at first and grows as it gets closer to its end of life. A piece-wise linear RUL target function was proposed in [3, 4], in order to better reflect the changes in Remaining Useful Life over time. This function sets a maximum RUL and then begins linear degradation at a specific usage level as shown in the Fig. 4. We set the maximum limit as 130 time cycles for all the engines.

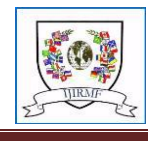

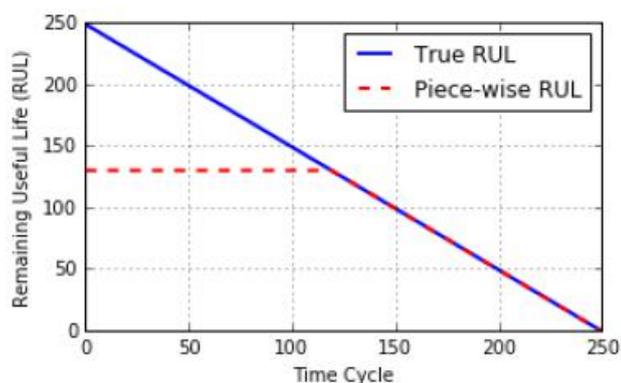

*Fig. 4: Piece-wise RUL of the Data Set (Piece-wise maximum RUL is 130-time cycles)*

### 4.2. Data Analysis

The Fig. 5 shows the sensor readings for one of the engines over the course of its lifetime. The sensor values are displayed on the Y-axis, while the power cycles are represented on the X-axis. From the graph, it can be observed that sensors such as Sensor 3, Sensor 4, Sensor 8, Sensor 9, Sensor 13, Sensor 19, Sensor 21, and Sensor 22 maintain constant readings throughout their lifecycle. Since these constant sensor readings do not contribute any predictive value, these features can be safely removed from the dataset.

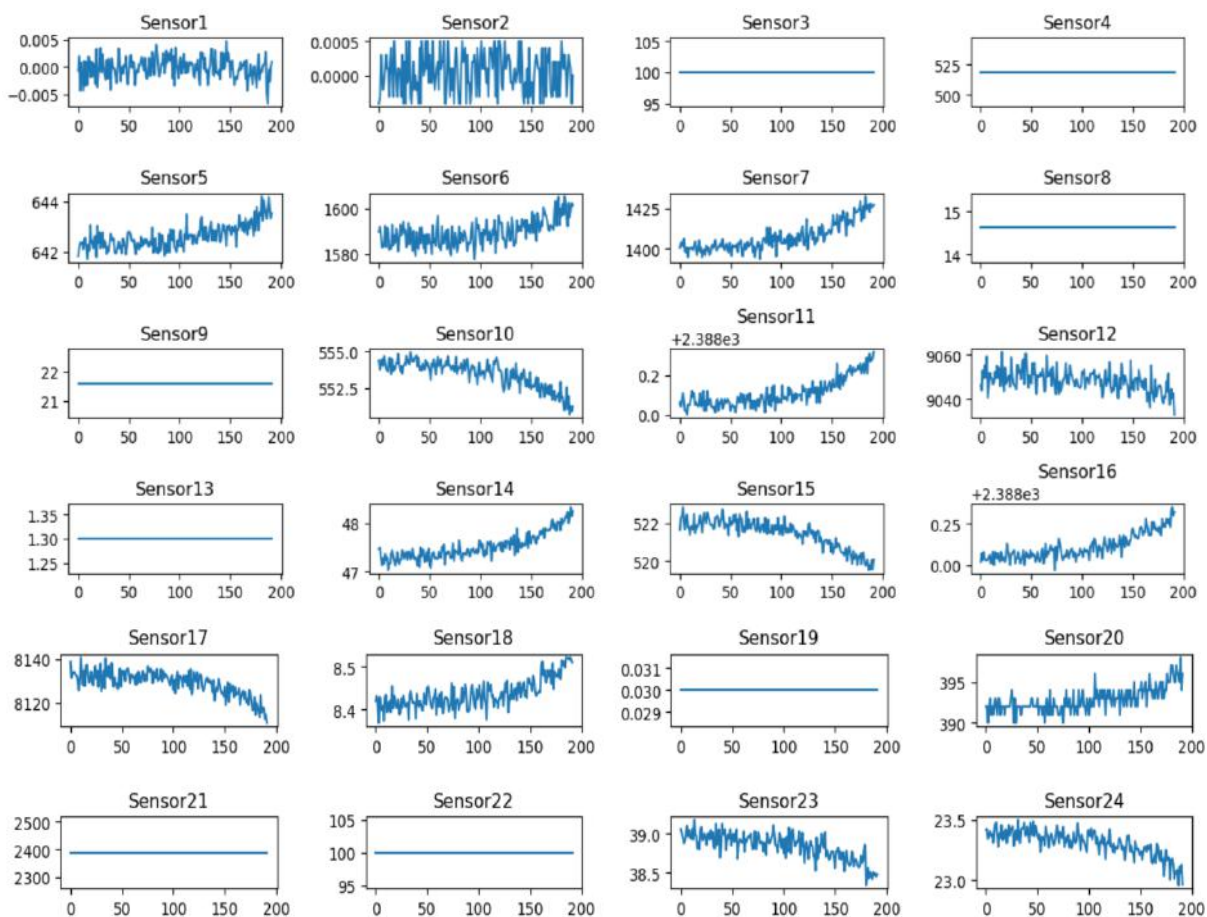

*Fig 5: Time series sensor data of one of the Engines*

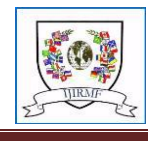

The distribution of each sensor data in the dataset varies significantly. As per Fig. 6, some sensors, such as 1, 5, and 6, exhibit a normal distribution and the remaining sensors display skewed distributions, either to the right or left. This diverse range of data distributions necessitates tailored pre-processing techniques to ensure optimal model performance.

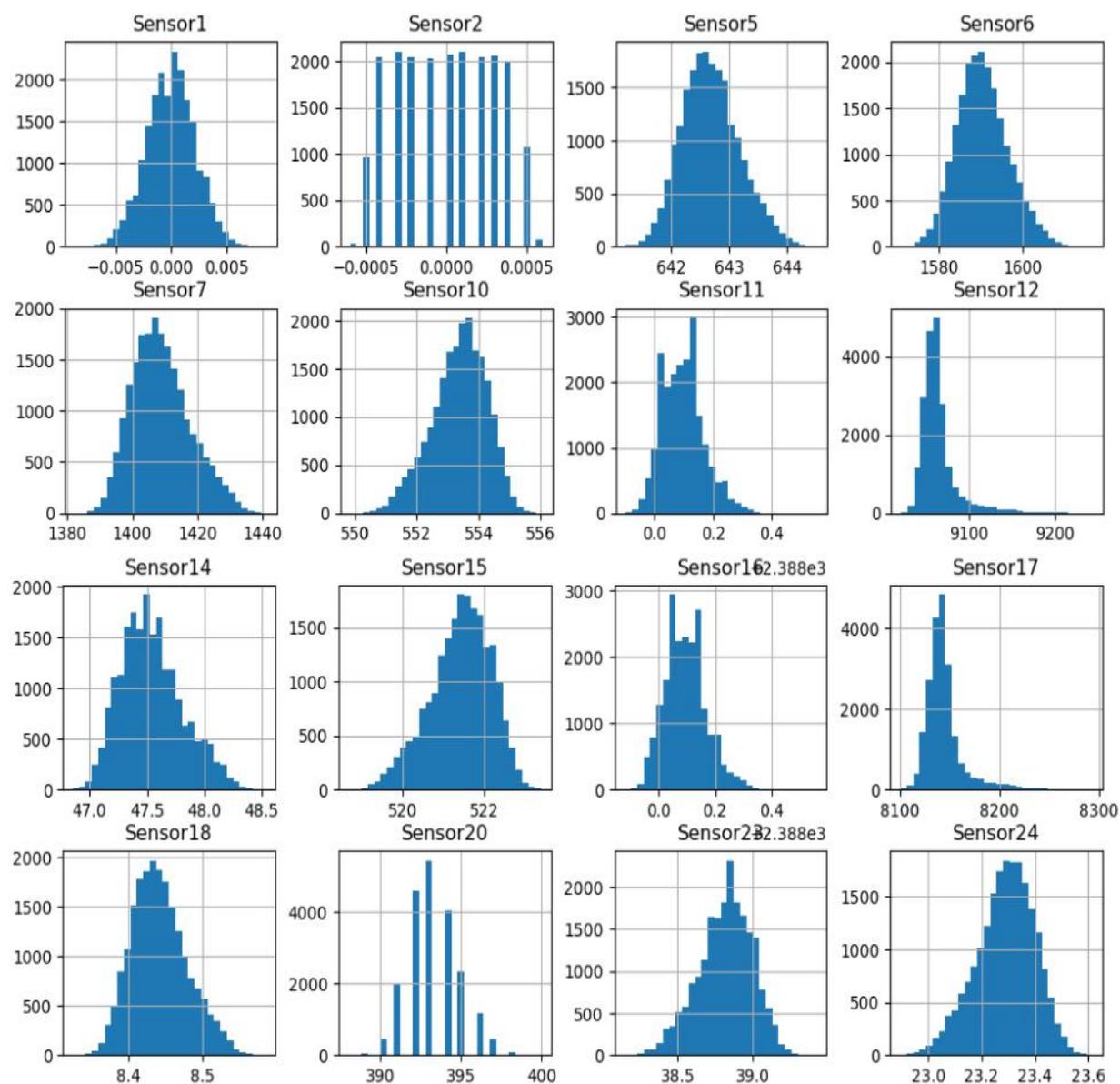

*Fig. 6: Frequency Distribution Plot for Sensor Data*

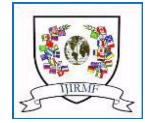

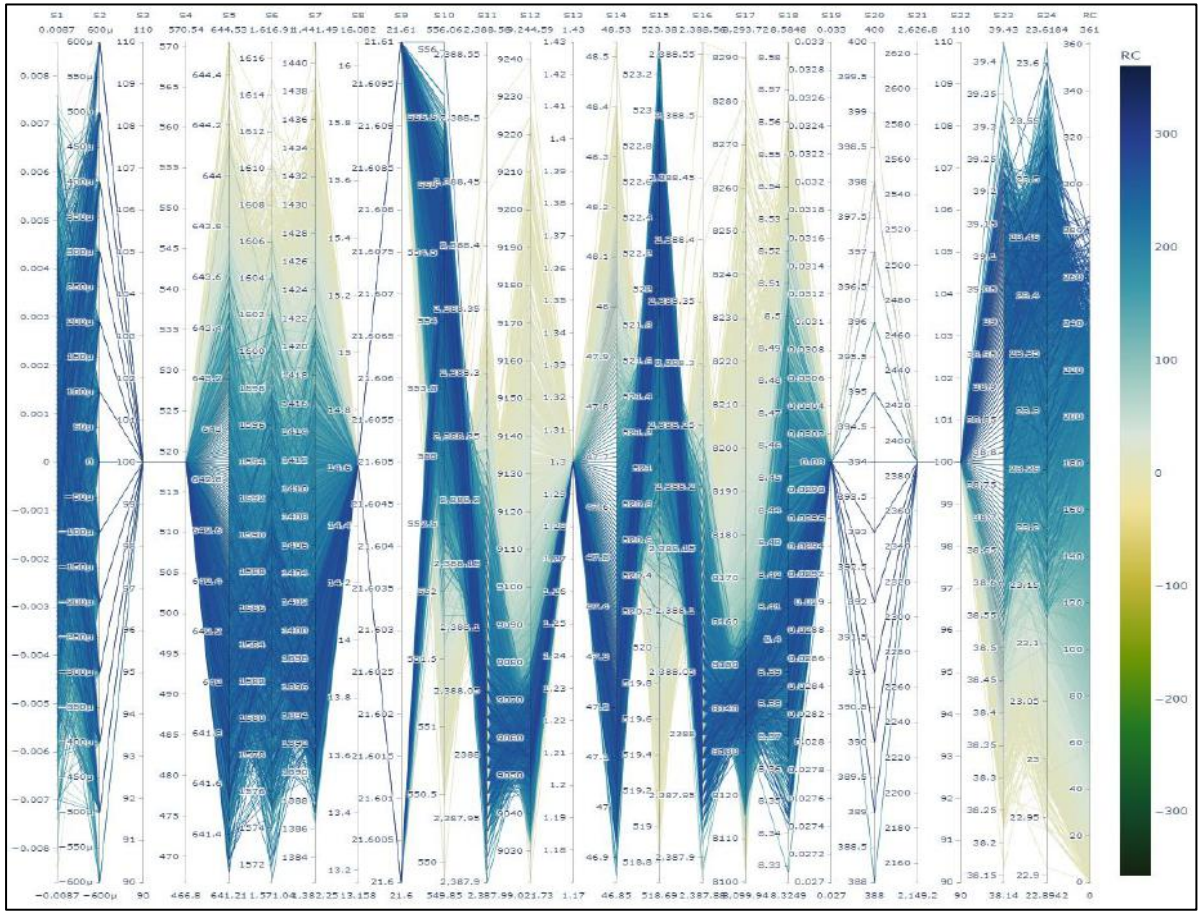

*Fig. 7: Parallel Coordinates plot for CMAPSS Dataset*

Parallel Plot [11] is a type of data visualization used to explore and analyse multi-dimensional data. Parallel plot applied to the dataset is displayed in Fig. 7, where dark blue lines represent observations with more remaining cycles, while yellow lines indicate lower remaining cycles. The sensor values tend to be more consistent and grouped together in the early cycles, which are represented by dark blue lines and suggest stable operation. Sensor readings begin to differ as the engines get closer to the middle of their lives; this is a sign of both performance discrepancies and the onset of defects. The sensor values exhibit a substantial divergence, suggesting wear and degradation, as the sensor approaches its end of life, indicated by yellow lines.

The sensor data for all engines is illustrated in the Fig. 8. The green dots in this graphic show the exponential moving average of the sensor data for one engine throughout the course of its lifetime, while the red points show the points at which each engine fails. With the exception of Sensors 1 and 2, it is clear from the figure that the majority of sensors are able to differentiate between the engine's normal operating circumstances and failure scenarios. As a result, the characteristics associated with Sensors 1 and 2 can be disregarded for additional examination. Furthermore, some sensor readings are positive correlation with degradation and while others show a negatively correlation.

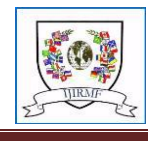

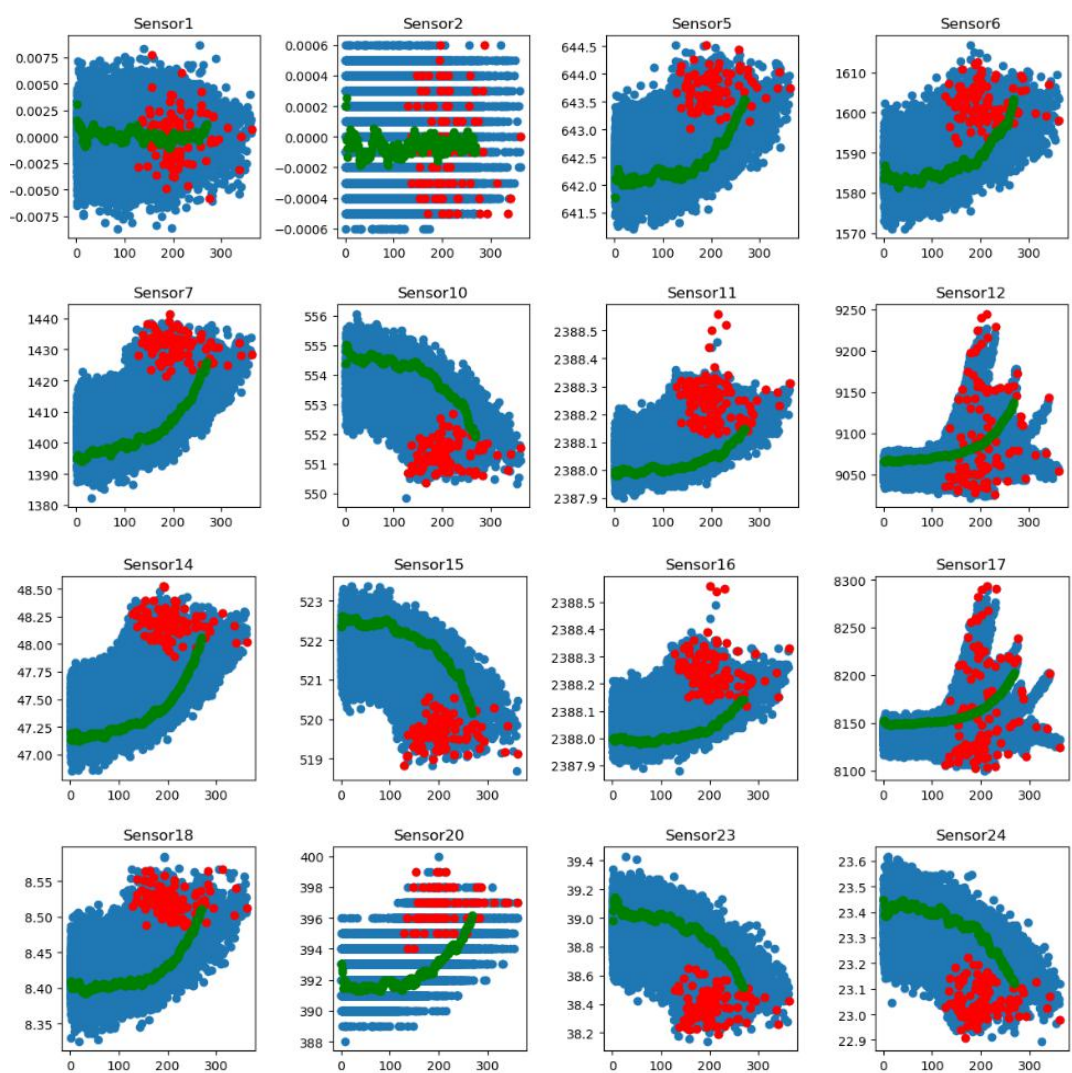

*Fig. 8: Scatter plot representation of sensor data of all engines*

### 4.3. Data Pre-processing

Data preprocessing is essential before modeling as it enhances data quality, ensuring that the input data is accurate, consistent, and relevant.  The techniques such as cleaning and standardization rectify errors and inconsistencies, leading to more reliable models.

### 4.3.1. Data Filtering

The sensor data in the dataset are noisy and sporadic, necessitating the application of smoothing filters to improve data quality. Two widely used smoothing techniques such as Simple Moving Average (SMA) and Exponential Moving Average (EMA), were applied with various weights to determine the most effective method. Upon evaluation, EMA with an alpha value of 0.1 visually outperformed other configurations. Consequently, EMA with this alpha value was selected and applied to the sensor data, resulting in a smoother and more reliable data. Fig. 9 shows a visual comparison of raw sensor data and the exponential mean of sensor data with alpha = 0.1. In the raw data, the data points are far more scattered than their corresponding exponential mean. Therefore, the modified data may provide better results for the model than using the raw sensor data directly.

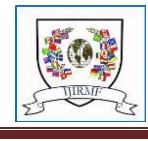

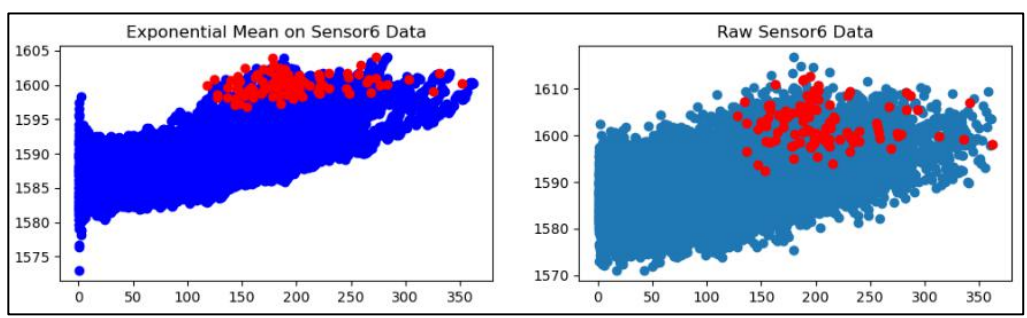

*Fig. 9: Comparison between Raw sensor data and exponential moving averaged sensor data*

### 4.3.2. Data Standardization

Since the value range is substantially different in different variables, it can be difficult to find the optimal point for the cost function. Therefore, the training and testing datasets need to be normalized. There are two widely used methods for normalization, which are Z-scores as specified in Eq. (4) and min-max-scale as specified in Eq. (5). Both methods are applied, and the one with the best evaluation result is selected.

$$x' = \frac{x - \text{mean}(x)}{\text{std}(x)} \qquad (4)$$

$$x' = \frac{x - x_{\min}}{x_{\max} - x_{\min}} \qquad (5)$$

### 4.4. Feature Engineering

In the dataset, some features exhibit constant values throughout all observations, resulting in zero variance. These features are removed in the previously discussed data analysis section, as they provide no meaningful insight into the relationship between the input variables and the target variable. In addition to that, feature transformation can be applied to further optimize the dataset. By removing or transforming such features, the model's performance can be significantly improved.

### 4.4.1. Principal Component Analysis

As part of feature engineering, Principal Component Analysis (PCA) was applied to reduce the dimensionality of the high-dimensional sensor data. PCA projects the data onto orthogonal axes, retaining the most significant features. The cumulative explained variance ratio, shown in Fig. 10, indicates that the first two components capture around 70% of the variance, while 12 components account for nearly 99%. Therefore, the data dimension can be reduced from 24 to 12 with minimal loss of information.

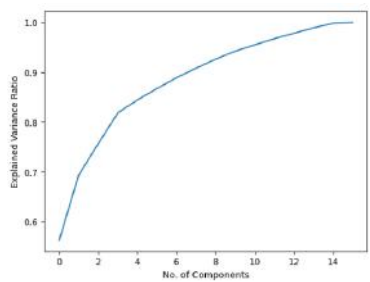

*Fig. 10: Explained Variance Ratio of Principal Components*

The first, second, and third principal components are studied extensively as they represent 75% of the dataset's variance. Fig. 11 represents the scatter plot for the complete dataset, where green dots represent the starting points of the engine cycles, and orange dots represent the failure points. It becomes evident from the scatter plot that thresholding the first principal component can effectively determine the failure point of all engines, irrespective of the other principal components.

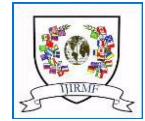

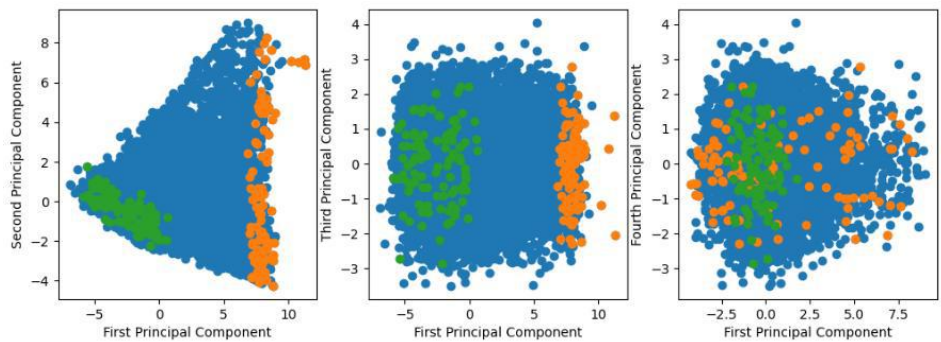

*Fig. 11: Correlation plot of first 3 principal components*

### 4.4.2. Multi collinearity Analysis

When working with multiple numerical features in a dataset, it is essential to analyse the correlation between these features. Highly correlated features can introduce redundancy, biasing the model and ultimately reducing the accuracy of predictions. When two or more features provide similar information, the model may overemphasize their contributions, leading to overfitting or skewed results. From the heat map shown in Fig. 12, it is evident that lot of features are highly correlated and it can lead to multi collinearity problem. Multicollinearity refers to the situation where two or more features in a dataset are highly correlated, which can negatively impact the performance of machine learning models. Therefore, it is crucial to select features that contribute the most to the model's performance while minimizing redundancy.

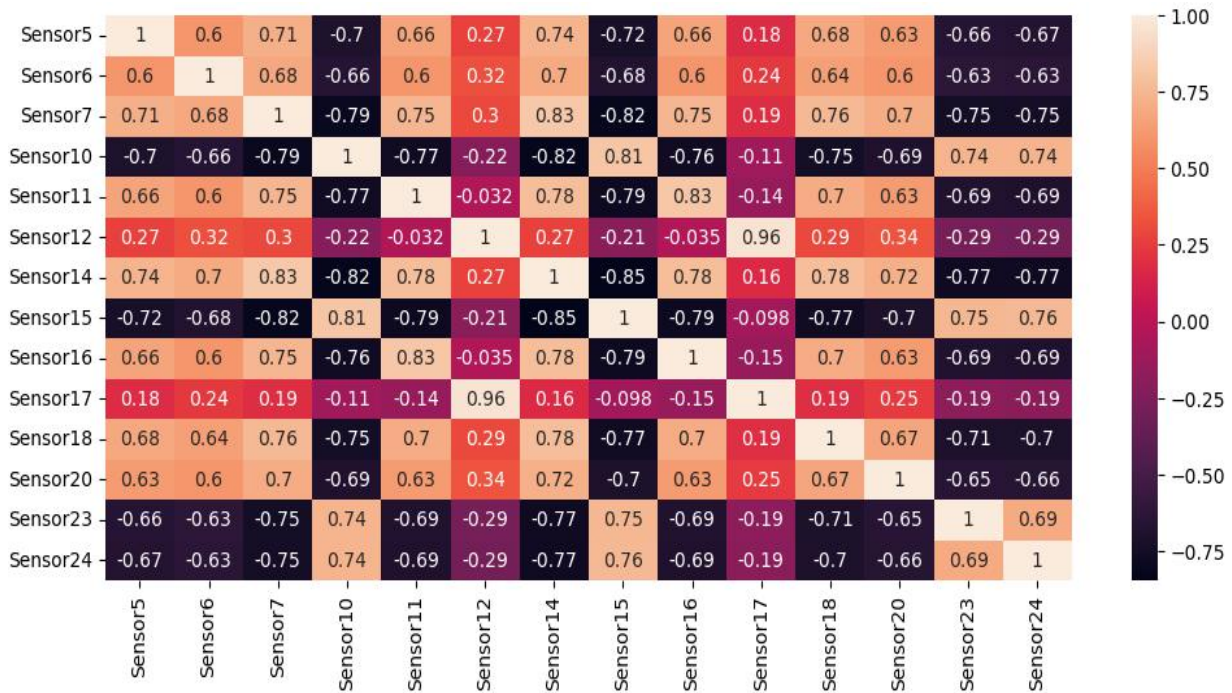

*Fig. 12: Heat map of the features in CMAPSS dataset*

### 4.4.3. Feature Selection

From the Fig. 13, it is evident that the first principal component can be segregated to indicate failure, which helps the model in tuning the remaining useful life (RUL). Therefore, the first principal component is included as an additional feature for model development. This inclusion enhances the model's ability to predict RUL more accurately by leveraging the significant variance captured by the first principal component.

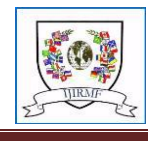

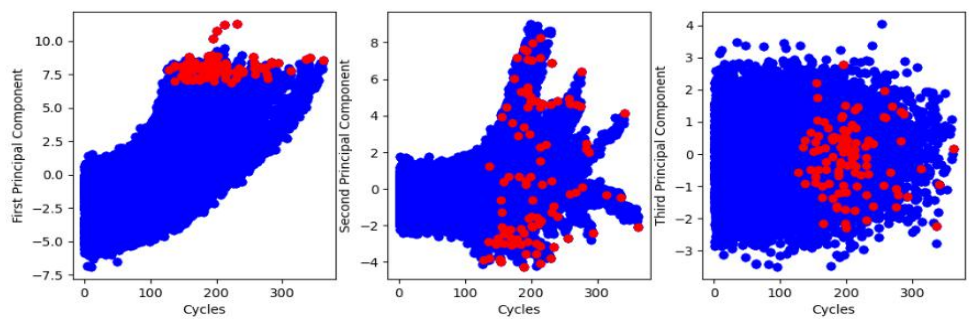

*Fig. 13: Principal Components Vs Cycles*

To tackle the issue of multicollinearity, as discussed in the previous section, we applied the Select K Best algorithm. This algorithm ranks all the features according to a specified statistical criterion and selects the top K features that are most significant for predicting the target variable. Fig. 14 clearly shows that the logarithmic F-scores of Sensor1 and Sensor2 are significantly low, indicating their minimal importance. Therefore, these sensors can be removed due to their limited contribution.

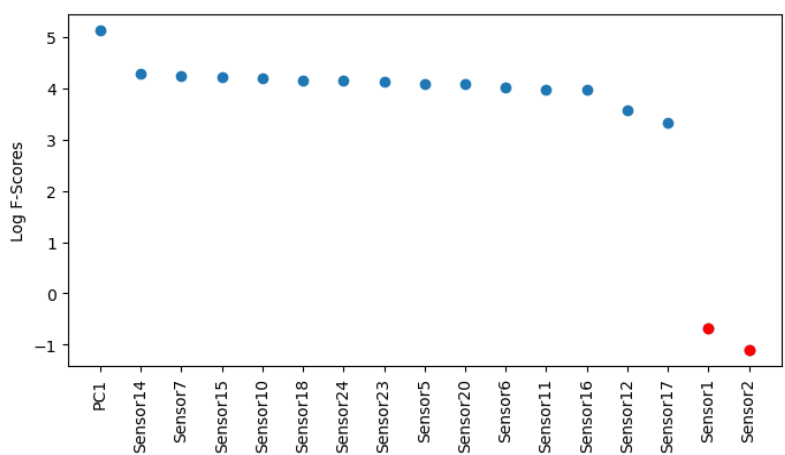

*Fig. 14: Features ranked based on F-scores by Select K Best Algorithm*

## 5. Experiments and Results

In this section, we have performed extensive experiments for comparison of the proposed CNN LSTM based deep learning model with traditional regression algorithms such as Linear Regression, Random Forest [6], and state-of-the-art algorithms, including Multi-layer Perceptron (MLP) [9], XGBoost [7, 8] and LSTM on the CMAPSS data set. The hyper parameters of all the techniques, are chosen using standard 5-fold cross-validation procedure, where we tune their parameter values for training these models and choose their final values that give the best results.

The hyperparameter tuning for the Random Forest, XG boost, MLP are performed as shown in the Table 1 and Table 2.

*Table 1: Hyperparameter Tuning for Random Forest*

| Hyperparameter | Description | List of Values | Best Estimator |
|---|---|---|---|
| n_estimators | No. of Trees | [100,200,300] | **300** |
| max_depth | Maximum depth of trees | [6,8,10,12,14] | **6** |
| min_samples_leaf | Minimum of samples for leaf | [4,6,8,10] | **4** |
| ccp_alpha | Tree pruning factor | [0,1,2] | **0** |

*Table 2: Hyperparameter Tuning for XG Boost*

| Hyperparameter | Description | List of Values | Best Estimator |
|---|---|---|---|
| learning_rate | Learning Rate | [0.05,0.1,0.2] | **0.1** |

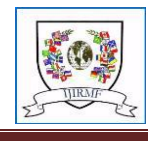

| n_estimators | No. of Trees | [70,100,200,300] | **70** |
|---|---|---|---|
| max_depth | Maximum depth of trees | [3,4,5] | **3** |
| min_child_weight | Minimum sum of instance weights | [70,100,150,200] | **200** |

*Table 3: Hyperparameter Tuning for MLP*

| *Hyperparameter* | *Description* | *List of Values* | *Best Estimator* |
|---|---|---|---|
| learning_rate | Learning Rate | [0.05,0.1,0.2] | **0.1** |
| n_layers | No. of layers | [3,4,5,6] | **5** |
| layer_sizes | No. of neuron in layers | [8.16.64.32.8, 16.32.64.32.16, 32.64.64.32.16] | **32.64.64.32.16** |

The performance of the previously discussed machine learning algorithms are compared with the performance of the proposed model and the results are shown in Table 4. From the table, it is evident that the proposed hybrid CNN-LSTM model demonstrates the better RMSE and offering a superior $R^2$ score compared to the other methods.

*Table 4: Performance comparison on C-MAPSS dataset*

| *Model* | *RMSE* | $R^2$ |
|---|---|---|
| Linear Regression | 43.18 | 0.46 |
| Random Forest | 6.68 | 0.42 |
| XG Boost | 17.35 | 0.65 |
| MLP | 4.51 | 0.52 |
| LSTM | 15.93 | 0.75 |
| CNN LSTM | **13.34** | **0.86** |

## 6. Conclusion and Future work :

We proposed CNN LSTM based deep learning approach for RUL estimation and we showed its benefits by taking sequence information when estimating RUL. Our experiments on C-MAPSS dataset showed that our proposed model outperforms other approaches and gives the best performance in RUL estimation. In addition to that, the work involves the entire lifecycle of predictive maintenance, starting with data collection, followed by comprehensive data preparation and pre-processing stages to ensure data quality and consistency. Feature scaling, Principal Component Analysis (PCA), and feature selection techniques were applied to refine the dataset and identify the most relevant features for modeling. Multiple algorithms, including Linear Regression, Random Forest, XG-Boost, MLP, and LSTM, were developed and evaluated using RMSE and R² metrics. Future work will focus on extending this experiment to other RUL estimation datasets to validate the robustness of the proposed model across different scenarios. A notable challenge with the proposed model is its computational complexity, which impacts its suitability for deployment in embedded devices. To address this, optimization strategies will be explored to increase the model's computation speed, making it more efficient and feasible for real-time applications on low-power embedded systems.

## 7. Acknowledgements

The authors would like to express their gratitude and appreciation to their team members, Abhay Sharma, Anchal Sekhri, Sana Zehra and Khunwana Zeno, for their invaluable inputs and support during the course of a project related to this research work.

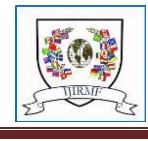

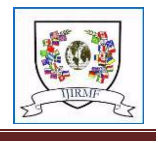